\documentclass{tlp}

\usepackage{amsmath}
\usepackage{graphicx}
\usepackage{multirow}

\usepackage{siunitx}
\usepackage{todonotes}
\usepackage{xspace}
\usepackage{subcaption}
\usepackage{amsfonts}
\usepackage{xcolor}
\usepackage{ulem}
\usepackage[hidelinks]{hyperref}
\usepackage{marvosym}
\usepackage{makecell}

\newcommand{\fodot}{FO($\cdot$)\xspace}
\newcommand{\new}[1]{{#1}}
\newcommand{\old}[1]{{#1}}
\usepackage[most]{tcolorbox}     

\newcommand{\blockquote}[1]{
\begin{tcolorbox}[boxrule=0pt,frame hidden,sharp corners,enhanced,borderline west={2pt}{0pt}{black!25!white},grow to left by=-10mm, grow to right by=-10mm] ``#1'' \end{tcolorbox}
}

\begin{document}

\lefttitle{Simon Vandevelde et al.}

\jnlPage{1}{8}
\jnlDoiYr{2021}
\doival{10.1017/xxxxx}

\title[Knowledge-Based Support for Adhesive Selection]{Knowledge-Based Support for Adhesive Selection: Will it Stick?\thanks{This research received funding from the Flemish Government under the ``Onderzoeksprogramma Artifici\"ele Intelligentie (AI) Vlaanderen'' programme and Flanders Make vzw.}}

\begin{authgrp}
\author{\sn{Simon} \gn{Vandevelde}}
\affiliation{KU Leuven, De Nayer Campus, Dept. Of Computer Science,\\J.-P. De Nayerlaan 5, 2560 Sint-Katelijne-Waver, Belgium}
\affiliation{Leuven.AI --- KU Leuven institute for AI, B-3000, Leuven, Belgium}
\affiliation{Flanders Make --- DTAI-FET}
\email{s.vandevelde@kuleuven.be}
\author{\sn{Jeroen} \gn{Jordens}}
\affiliation{Flanders Make, Oude Diestersebaan 133, 3920 Lommel, Belgium}
\email{jeroen.jordens@flandersmake.be}
\author{\sn{Bart} \gn{Van Doninck}}
\affiliation{Flanders Make, Oude Diestersebaan 133, 3920 Lommel, Belgium}
\email{bart.vandoninck@flandersmake.be}
\author{\sn{Maarten} \gn{Witters}}
\affiliation{Flanders Make, Oude Diestersebaan 133, 3920 Lommel, Belgium}
\email{maarten.witters@flandersmake.be}
\author{\sn{Joost} \gn{Vennekens}}
\affiliation{KU Leuven, De Nayer Campus, Dept. Of Computer Science,\\J.-P. De Nayerlaan 5, 2560 Sint-Katelijne-Waver, Belgium}
\affiliation{Leuven.AI --- KU Leuven institute for AI, B-3000, Leuven, Belgium}
\affiliation{Flanders Make --- DTAI-FET}
\email{joost.vennekens@kuleuven.be}
\end{authgrp}

\history{\sub{xx xx xxxx;} \rev{xx xx xxxx;} \acc{xx xx xxxx}}

\maketitle

\begin{abstract}
  As the popularity of adhesive joints in industry increases, so does the need for tools to support the process of selecting a suitable adhesive.
  While some such tools already exist, they are either too limited in scope, or offer too little flexibility in use.
  This work presents a more advanced tool, that was developed together with a team of adhesive experts.
  We first extract the experts' knowledge about this domain and formalize it in a Knowledge Base (KB).
  The IDP-Z3 reasoning system can then be used to derive the necessary functionality from this KB.
  Together with a user-friendly interactive interface, this creates an easy-to-use tool capable of assisting the adhesive experts.
  To validate our approach, we performed user testing in the form of qualitative interviews.
  The experts are very positive about the tool, stating that, among others, it will help save time and find more suitable adhesives.

\end{abstract}


\section{Introduction}

The Flanders Make Joining \& Materials Lab (FM JML) is specialized in adhesive bonding.
They support companies in selecting the most appropriate adhesive for a specific use case, by accounting for characteristics such as strength, temperature resistances, adhesive durability, and more.
Currently, this is done manually by one of the \textit{adhesive experts} that work at the lab.
Selecting a suitable adhesive is a time-consuming and labor intensive task, due to the large number of adhesives available on the market, each with extensive data sheets.
Currently, the experts do not use any supporting tools to help them perform the selection, because the current generation of tools does not meet their requirements.


This paper describes our work on a logic-based tool which supports the experts in the selection process.
It is structured as follows.
We start by describing the process of selecting an adhesive and the state-of-the-art tools in Section~\ref{case}, and elaborate on the logical system used in this work in Section~\ref{kbp}.
Next, we present our Adhesive Selector Tool in Section~\ref{tool}, where we discuss the process of Knowledge Acquisition, how the system handles unknown parameter values, and how the experts interface with the knowledge.
We share the results of our preliminary three-fold validation in Section~\ref{validation} and the results of our comprehensive user study in Section~\ref{s:userstudies}.
Finally, we describe our lessons learned in Section~\ref{lessons} and conclude in Section~\ref{conclusion}.

This paper is an extended version of~\cite[]{Vandevelde2022}, as presented at the LPNMR22 conference.
Its main addition is a user-study with the target users of the JML lab, in the form of semi-structured interviews.
Additionally, we elaborated on some sections and updated the text with various minor improvements.

\section{Adhesive Selection and Current Tools}\label{case}

As there is no universally applicable adhesive, the selection of an adhesive is an important process.
There are many factors that influence the choice of an adhesive: structural requirements such as bonding strength and maximum elongation, environmental factors such as temperature and humidity, economic factors, and more.
Due to the complexity of the problem, there is quite a potential for tools that support this selection process.
Yet, \cite{EWENJC20103} concludes that ``there is a severe shortage of selection software, which is perplexing especially when the task of adhesive selection is so important.''

Currently, when tasked with a use case, the experts work in two steps.
First, they try to identify requirements, such as temperature ranges or values for parameters like minimum strength.
Based on this list of requirements, they perform an \new{initial} selection by manually looking through various data sheets while keeping track of which adhesives are suitable.
In the second step, the\old{se} \new{inititally chosen} adhesives are put to the test by performing real-life experiments in FM's lab, to ensure suitability.
However, this testing step is costly and time-consuming, so it is important that the initial selection is as precise as possible.
While there are tools available for this process, the FM experts do not use them because they are either too simplistic, or not sufficiently flexible.


The most straightforward selection tools are websites offering simple interfaces\footnote{such as \url{www.adhesivestoolkit.com} and \url{www.adhesives.org}}~\cite[]{da2018handbook}.
Based on a series of questions, they provide advice to support selection.
However, they still require the expert to look up and process the information themselves.

There are also a number of expert systems to be found in the literature \cite[]{kannanexpert,rb1995expert,su1993knowledge,moseley1992development,meyler1993design,lees1993pal,lammel2002software}.
Here, domain knowledge is captured and formalized in the form of rules, which can be used for adhesive selection by forward chaining and often also for generating explanations by backward chaining.
However, these systems have a number of downsides: they are low in both interpretability and maintainability by the expert, often not all required knowledge can be expressed, and they generally only contain a low number of adhesives or substrates.
Finally, forward and backward chaining are not capable of providing all the functionality the expert needs.
For instance, a situation might arise in which an adhesive is already pre-defined (e.g., left-over from a previous gluing operation), and the selection of a second substrate is required.
While this selection requires the same knowledge, the expert tools are not capable of performing this operation.

\section{Knowledge Base Paradigm}\label{kbp}

The core idea in the Knowledge Base Paradigm (KBP)~\cite[]{KBP} is to represent knowledge in a purely declarative way, independently of how it is to be applied.
Knowledge is stored in a Knowledge Base (KB) and can be put to use via a multitude of inference tasks.
In this way, the approach stimulates \textit{knowledge reuse}, as multiple inference tasks can be used to solve multiple problems with the same knowledge.


\subsection{IDP}
The IDP (Imperative Declarative Programming) system~\cite[]{IDP} is an implementation of the KBP.
The knowledge in the KB is represented in a rich extension of First Order Logic (FOL), called \fodot (pronounced \textit{FO-dot}).
It extends FOL with types, aggregates, inductive definitions and more. 
\fodot is an expressive and flexible knowledge representation language, capable of modeling complex domains.
The knowledge in a KB is structured in three kinds of blocks: \textit{vocabularies}, \textit{structures} and \textit{theories}.

A \textit{vocabulary} specifies  a set of symbols.
A symbol is either a type, predicate, or a function.
A type is either a standard type such as the set of real numbers $\mathbb{R}$, or the name of an application-specific type, such as $Adhesive$.
A predicate symbol either expresses a boolean or a \textit{relation} on one or more types, such as \textit{BondSealing()} or \textit{Available(Adhesive)}.
A function symbol represents a function from the Cartesian product $T_1 \times \ldots \times T_n$ of a number of types to a type $T_{n+1}$.
For example, the function $\mathit{BondStrength}: \mathit{Adhesive} \rightarrow \mathbb{R}$ maps each adhesive on its bond strength.

A \textit{(partial) structure} specifies an interpretation for some of the symbols of a given vocabulary.
A structure is \textit{total} if it specifies an interpretation for each symbol of the vocabulary.

A \textit{theory} contains a set of logical formulae in \fodot.

By itself, the KB cannot be executed: it is merely a ``bag of knowledge'', without information on how it should be used.
The latest version of the IDP system, IDP-Z3~\cite[]{idpz3}, supports many different inference tasks that can be applied to this knowledge.
We will briefly go over the inference tasks that are relevant: \textbf{propagation}, \textbf{model expansion}, \textbf{optimization} and \textbf{explanation}.
Given a partial interpretation $\mathcal{I}$ for the vocabulary of a theory $\mathit{T}$, \textbf{propagation} derives the consequences of $\mathcal{I}$ according to $\mathit{T}$, resulting in a more precise partial interpretation $\mathcal{I'}$.
\textbf{Model expansion} extends a partial structure $\mathcal{I}$ to a complete interpretation $I$ that satisfies the theory T ($I \models T$).
\textbf{Optimization} is similar to model expansion, but looks for the model with the lowest/highest value for a given term.
Finally, \textbf{explanation} will, given a structure $\mathcal{I}$ which does not satisfy the theory T ($\mathcal{I} \not\models T$), find minimal subsets of the interpretations in $\mathcal{I}$ which together explain why $\mathcal{I}$ does not satisfy the theory.

\subsection{Interactive Consultant}

The Interactive Consultant~\cite[]{IC} is a graphical user interface for IDP-Z3, which aims at facilitating interaction between a user and the system.
It is a generic interface, in the sense that it is capable of generating a view for any syntactically correct KB.
In short, each symbol of the KB is represented using a \textit{symbol tile}, which allows users to set or inspect that symbol's value.
In this way, the GUI represents a partial structure to which a user can add and remove values.
Each time a value is added, removed or modified, IDP's propagation is performed and the interface is updated: symbols for which the value was propagated are updated accordingly, and for the other symbols the values that are no longer possible are removed.
In this way, a user is \textit{guided} towards a correct solution: they cannot enter a value that would make the partial structure represented by the current state of the GUI inconsistent with the theory.

At any point in time the user can ask for an explanation of a value that was derived by the system, e.g., when the user does not understand it or agree with it.
The system will then respond with the relevant formulas and user-made assignments that lead to the derived value.
In this sense, the tool is explainable, leading to more trust in the system.

A similar functionality is in place for the rare cases in which a user manages to reach an \textit{inconsistent} state, i.e., a set of assignments that can no longer be extended to a solution.
While the IC removes values that have become impossible, it cannot do so for variables belonging to unbounded integer or real domains.
For example, it is possible to input ``Min Temperature = 20'' followed by ``Max Temperature = 10'', as both of these variables can have any value between $-\infty$ to $+\infty$.
In the case of an inconsistency, the interface alerts the user and explains why no solutions are possible by showing the relevant design choices and laws.

The Interactive Consultant interface has already successfully been used in multiple applications in the past~\cite[]{Aerts2022, Deryck2019}.


\section{Adhesive Selector Tool}\label{tool}

This section outlines the creation and usage of the tool, and the main challenges that were faced in that process.
\subsection{Knowledge Acquisition}

The creation of the knowledge base is an important element in the development process of knowledge-based tools.
It requires performing knowledge acquisition, which is traditionally the most difficult step, as the knowledge about the problem domain needs to be extracted from the domain expert to be formalized by the knowledge engineer.
While knowledge acquisition comes in many shapes and forms, we applied the \textit{Knowledge Articulation} method~\cite[]{KnowledgeArticulation}.
The central principle of this method is to formalize knowledge in a common notation for both domain expert and knowledge engineer, so that both sides actively participate in the formalization process.

We started by organizing three knowledge articulation workshops, each lasting between three and four hours.
Each of these workshops was held with a group of domain experts.
While typically a single domain expert would suffice for knowledge extraction, having a group present can help as an initial form of knowledge validation, as the experts discuss their personal way of working amongst themselves, before coming to a consensus.
For the common notation we used Constraint Decision Model and Notation (cDMN)~\cite[]{cDMN}, an extension of the Decision Model and Notation (DMN) standard~\cite[]{DMN}.
DMN is a user-friendly, intuitive notation for simple decision logic.
Its main component is the \textit{decision table}, which allows to define the value of a number of ``output variables'' in terms of a number of ``input variables''.
Decision tables are structured together in a \textit{Decision Requirements Diagram} (DRD), which is a graph that provides an overview of the total decision model by showing the connections between input variables (ovals) and decisions (rectangles).
cDMN aims to increase the expressiveness of DMN (e.g., by adding constraints and quantification) while maintaining this user-friendliness.

\begin{figure}[h!]
    \centering
    \includegraphics[width=0.8\linewidth]{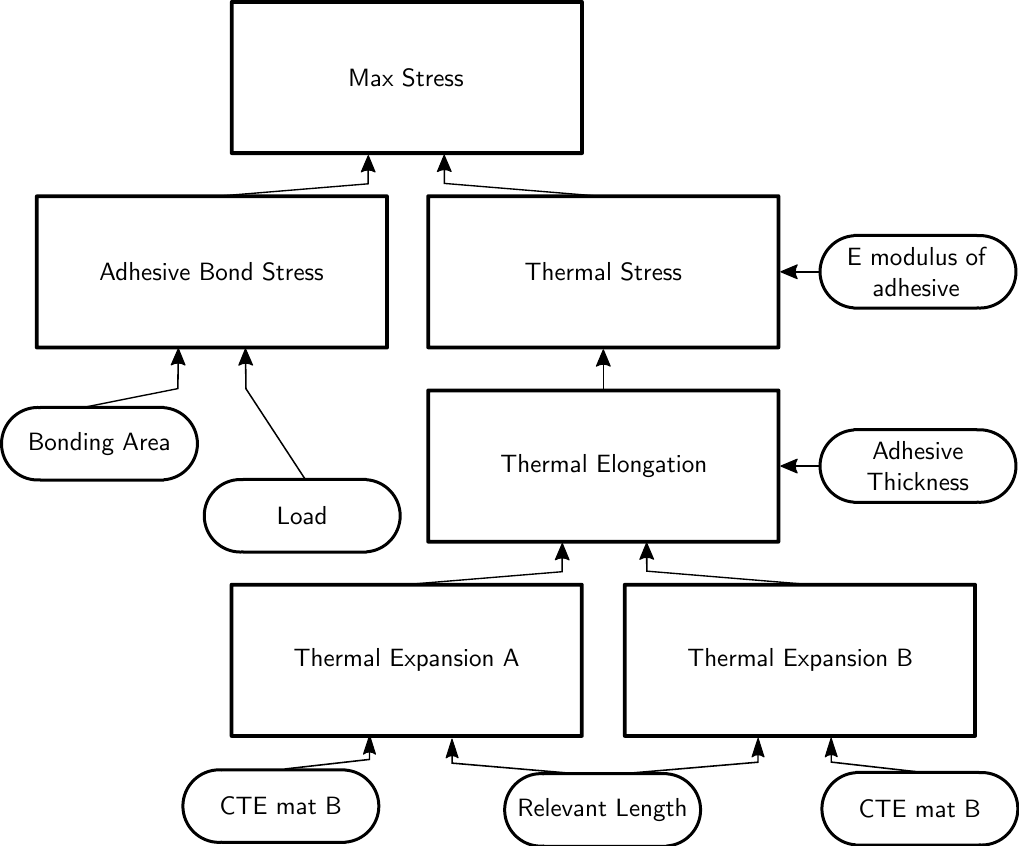}
    \caption{Snippet of created DRD.}
    \label{fig:drd}
\end{figure}

The first workshop consisted of identifying all relevant adhesive selection parameters and using them to create an initial DRD, of which a fragment is shown in Fig.~\ref{fig:drd}.
It is structured in a bottom-to-top way, similar to how the experts would reason: they start by calculating the thermal expansions, and then work their way up to the calculation of the maximum stress.

\begin{figure}[h!]
    \begin{subfigure}{\columnwidth}
        \footnotesize
        \centering
        \includegraphics[width=0.65\linewidth]{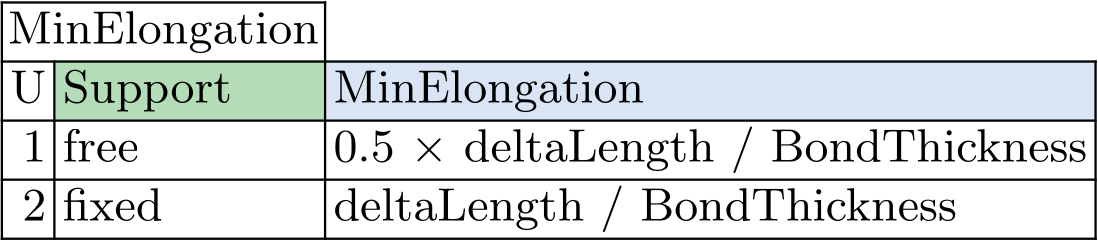}
        \caption{Calculation of MinElongation}
        \label{fig:decisiontable}
        \vspace{0.5em}
    \end{subfigure}
    \begin{subfigure}{\columnwidth}
        \centering
        \footnotesize
        \includegraphics[width=0.5\linewidth]{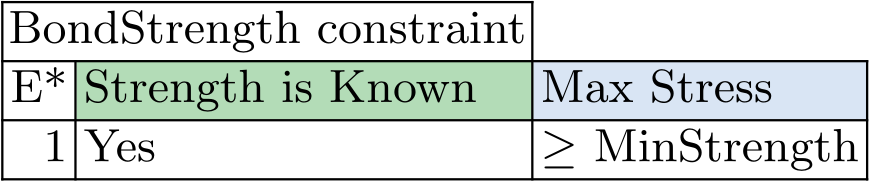}
        \caption{BondStrength constraint table.}
        \label{fig:constrainttable}
    \end{subfigure}
    \caption{Example cDMN tables}
    \label{fig:tables}
\end{figure}

During subsequent workshops, the rest of the model was fleshed out.
This consists of \textit{decision tables} and \textit{constraint tables}.
An example of such a decision table can be found in Fig.~\ref{fig:decisiontable}.
In such a table, the ``inputs'' (in green, left) define the ``outputs'' (in light blue, right).
Each row represents a decision rule, which \textit{fires} if the values of the input variables match the values listed in the row.
If a row fires, the value of the output is set accordingly.
E.g., if $\mathit{Support} = \mathit{fixed}$, then the \textit{MinElongation} is calculated as $\mathit{deltaLength} / \mathit{BondThickness}$.
The (U)nique hit policy of this table, indicated in the top left, means that the different rows must be mutually exclusive.

Fig.~\ref{fig:constrainttable} shows a constraint table (denoted by the \textit{E*} in the top-left corner).
In such a table, the output specifies a constraint that must hold if the input is satisfied.
In other words, this table states that \textit{if} the bond strength is known, \textit{then} \textit{Max Stress} should be higher than a minimum value.
This differs from decision tables in that it does not define a specific value, but rather constrains its possible values.

After these three initial workshops, the cDMN model was converted into an \fodot KB using the cDMN conversion tool.
Since then, multiple one-on-one workshops were held between the knowledge engineer (first author) and the primary domain expert (second author) to further fine-tune the KB.
Among other things, this included adding a list of adhesives, substrates, and their relevant parameter values, and further validating the knowledge.
In total, the current version of the KB contains information on 55 adhesives and 31 substrates.
For the adhesives, the KB contains 21 adhesive parameters, such as temperature resistances, strength and maximum elongation.
Similarly, it contains 11 parameters for the substrates, such as their water absorption and their solvent resistance.
These parameters are a mix of discrete and continuous: in total, 15 are continuous, and 17 are discrete.

\subsection{Unknown adhesive parameters}

One of the main challenges in the formalization of the KB was handling unknown adhesive data.
Indeed, often an adhesive's data sheet does not list all of its properties.
This raises the question of how the tool should deal with this: should the adhesive be excluded, or should it simply ignore the constraints that mention unknown properties?
Together with the experts we agreed on a third approach, in which we first look at the adhesive's family.
Each adhesive belongs to one of 18 families, for which often some indicative parameter values are known.
Whenever an adhesive's parameter is unknown, we use its family's value as an approximation.
If the family's value is also unknown, then the constraint is ignored.
This best corresponds to how the experts typically work.

This way of reasoning is formalized in the KB.
For example, the constraint that an adhesive should have a minimum required bonding strength is written as follows:
\begin{equation*}
    \begin{aligned}
        \forall p \in \mathit{param}: \mathit{Known}(\mathit{p}) \Leftrightarrow & (\mathit{KnownAdhesive}(\mathit{p})
        \lor \mathit{KnownFamily}(\mathit{p}))\\
    \end{aligned}
\end{equation*}
\begin{equation*}
    \begin{aligned}
        \mathit{KnownAdhesive}(\mathit{strength}) \Rightarrow \mathit{BondStrength} =& \mathit{StrengthAdhesive}(\mathit{Adhesive}).\\
        \neg\mathit{KnownAdhesive}(\mathit{strength}) \Rightarrow \mathit{BondStrength} =& \mathit{StrengthFamily}(\mathit{Family(Adhesive)}).\\
        \mathit{Known}(\mathit{strength}) \Rightarrow \mathit{BondStrength} \geq & \mathit{MinBondStrength}.
    \end{aligned}
\end{equation*}
with \textit{StrengthAdhesive} and \textit{StrengthFamily} representing respectively the specific adhesive's and its family's bonding strength.
This approach is used for all 21 adhesive parameters.

One caveat to this approach is that IDP-Z3 currently does not support partial functions, i.e., all functions must be totally defined.
To overcome this, we assign the value \textit{-1000} to unknown parameter values, and define that the value is only \textit{known} if it is different from this number.
We chose -1000 as there is no adhesive parameter for which it is a realistic value.
\begin{equation*}
    \begin{aligned}
        \forall p \in \mathit{param}: \mathit{KnownAdh(p)} \Leftrightarrow \mathit{StrengthAdhesive(Adhesive)} \neq -1000.\\
        \forall p \in \mathit{param}: \mathit{KnownFam(p)} \Leftrightarrow \mathit{StrengthFamily(Adhesive)} \neq -1000.\\
    \end{aligned}
\end{equation*}

%
%

\begin{figure}[!h]
\centering
    \begin{subfigure}{0.99\columnwidth}
        \centering
        \includegraphics[width=1\linewidth]{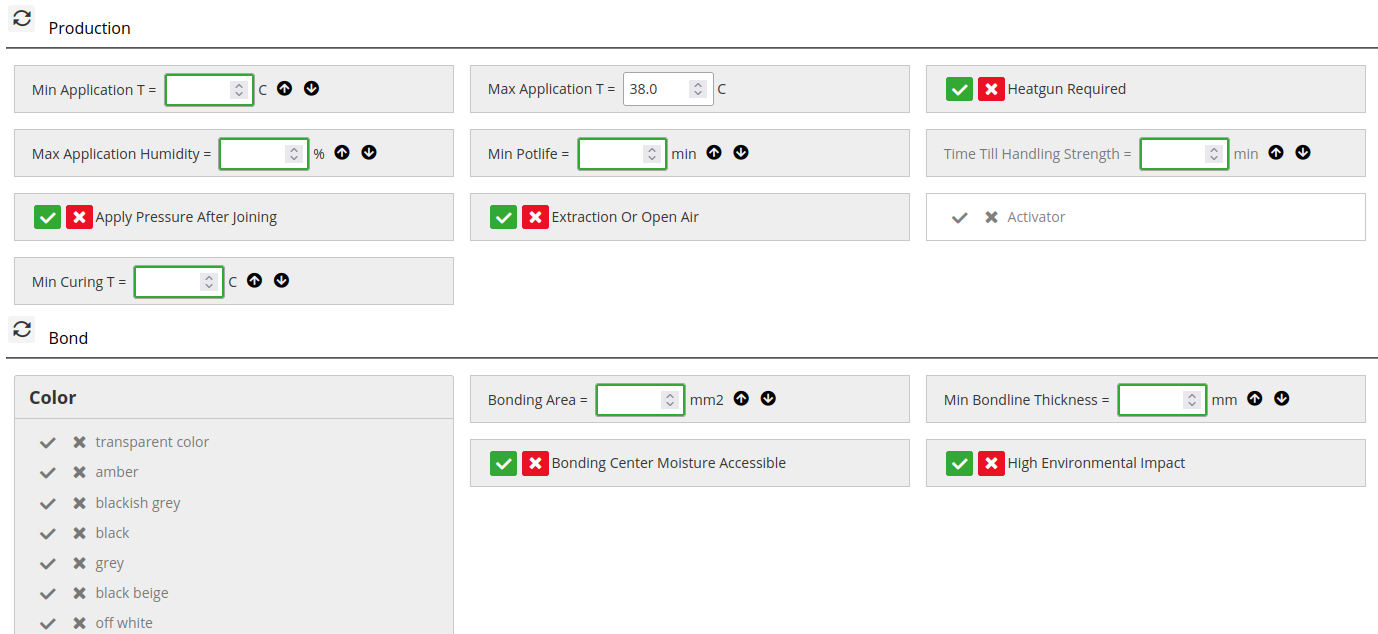}
        \caption{Some of the symbol tiles available in the interface, divided into two categories.}
        \label{fig:ic}
        \vspace{1em}
    \end{subfigure}
    \hspace{0.08\columnwidth}
    \begin{subfigure}{0.45\columnwidth}
        \centering
        \includegraphics[width=0.85\linewidth]{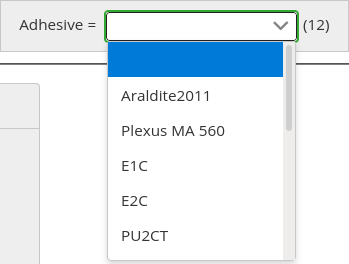}
        \caption{List of remaining suitable adhesives during selection.}
        \label{fig:adh}
    \end{subfigure}
    \begin{subfigure}{0.54\columnwidth}
        \centering
        \includegraphics[width=0.95\linewidth]{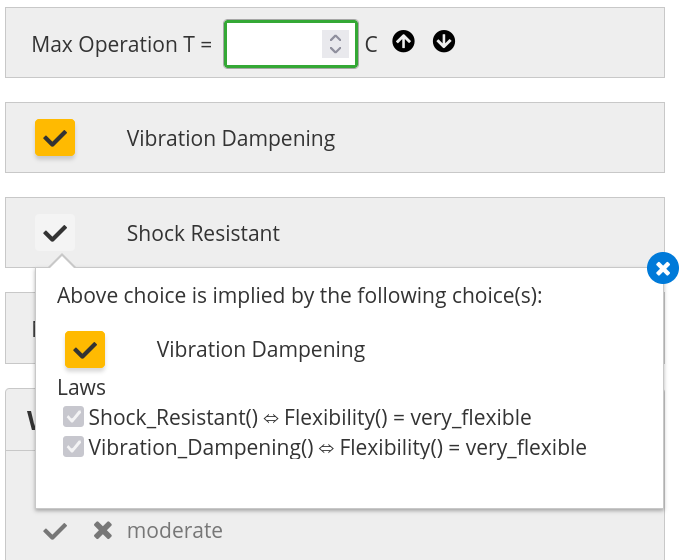}
        \caption{Example of explanation.}
        \label{fig:expl}
    \end{subfigure}
    \begin{subfigure}{\columnwidth}
        \centering
        \includegraphics[width=0.7\linewidth]{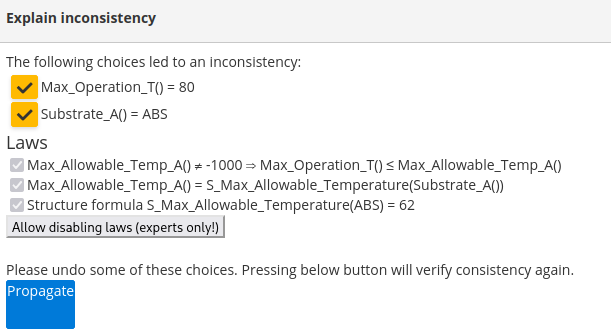}
        \caption{Inconsistency window.}
        \label{fig:inconsistency}
    \end{subfigure}
    \caption{Screenshots of the interface.}
\end{figure}

\subsection{Interface}

A crucial requirement of this application is the ability to interactively explore the search space.
To this end, our tool integrates the Interactive Consultant to facilitate interaction with the KB.
This interface makes use of several functionalities of the IDP system to make interactive exploration possible: the \textbf{propagation} inference algorithm is used to show the consequences of each choice, the \textbf{explain} inference is used to help the user understand why certain propagations were made, the \textbf{optimize} inference is used to compute the best adhesive that matches all of the choices made so far.

When using the interface, the user fills in symbol tiles, each representing a different symbol of the KB, and the system each time computes the consequences.
For example, Fig.~\ref{fig:ic} shows a segment of the interface in which a user set a maximum application temperature of $38^\circ C$ as a requirement.
To make it easier to navigate the symbol tiles, they are all divided in five \textit{categories}: Performance, Production, Bond, Substrate A and Substrate B.
In the top-right of the interface, the number of adhesives that remain feasible is shown: e.g., after setting the temperature constraint, that drops from 55 to 12, as shown in Fig.~\ref{fig:adh}.

The tool is also capable of generating two types of explanations.
Firstly, if the user does not understand why a certain value was propagated, they can click on that value to receive a clarification, as demonstrated in Fig.~\ref{fig:expl}.
Secondly, if the user manages to reach an inconsistent state, the tool will try to help resolving the issue by listing what is causing it.
For example, Fig.~\ref{fig:inconsistency} shows an inconsistency in which a substrate is selected that cannot handle the required operating temperature.

Besides generating a list of all the adhesives that meet certain requirements, the tool can also find the optimal adhesive according to a specific criterion, such as lowest price or highest strength.

\section{Preliminary Validation}\label{validation}

Initially, we performed three types of validation for this tool: a benchmark to measure the efficiency, a survey to measure the opinion of the adhesive experts and a discussion with the Flanders Make AI project lead.

\paragraph{Benchmark}
In an initial benchmark, an adhesive expert was tasked with finding a suitable adhesive for an industrial use case which the company received.
In total, it took the expert about three hours to find such an adhesive, after delving through multiple data sheets.
We then used our tool for the same use case, and were able to find the exact same adhesive within three minutes.
Interestingly, the reasoning of the tool closely mimicked that of the expert: for example, they both excluded specific families for the same reasons.

\paragraph{Survey}
After a demonstration of the tool to four adhesive experts, we asked them to fill out a short quantitative survey to better gauge their opinion.
Their answers can be summarized as follows.
\begin{itemize}
    \item The experts find the tool most useful for finding an initial list of adhesives to start performance testing with.
    \item The tool will be most useful for newer, less knowledgeable members of the lab. They can use the tool to learn more about the specifics of adhesive selection, or to verify if their result is correct.
    \item However, it is also useful for senior experts as they can discover adhesives which they have not yet used before.
\end{itemize}
The main criticism of the tool given by the experts is that more adhesives should be added, to make the selection more complete.

\paragraph{Project Lead Discussion}
As part of a discussion with Flanders Make's project lead, who oversees multiple AI-related projects, they outlined their perception of our tool.
They see many advantages.
Firstly, as there is not much data available on the process of adhesive selection (e.g., previous use cases and the selected adhesives), and data generation is quite expensive, data-based approaches are not feasible.
Therefore, building a tool based on a formalization of the knowledge they already have is very interesting.
Secondly, by ``storing'' the expert knowledge formally in a KB they can retain this information, even when experts leave the company.
Thirdly, having a formal representation also makes the selection process more uniform across different experts, who typically use different heuristics or rules-of-thumb.
Lastly, they indicated that there is trust in the system, because the knowledge it contains is tangible.
This makes it more likely that the experts will agree with the outcome of the tool.

The project lead also expressed that there is potential to maintain and extend this tool themselves, which would be a significant advantage compared to their other AI systems.
However, we currently have not yet focused on this aspect.

\section{User Study}\label{s:userstudies}

On top of the preliminary validation presented in Section~\ref{validation}, we performed a user study with members of the Flanders Make JML group.
This study distinguishes itself from the preliminary validation in two ways: (1) all experts actually got to use the tool themselves and (2) it is a qualitative study instead of quantitative w.r.t.\ the input of the experts.
Our main motivation for this study is to thoroughly validate the Adhesive Selector, with the following goals in mind:
\begin{enumerate}
    \item Gauge the tool's effectiveness in a real-life setting.
    \item Observe how new users interact with the tool.
    \item Get feedback on the different aspects of the tool (interactivity, interface, explanations, ...).
\end{enumerate}

For this validation, we asked the FM JML members to work out two real-life use cases, after which we performed one-on-one semi-structured interviews (SSI).
In the following subsections we first elaborate on our methodology, then discuss the results of our interviews followed by describing the limitations of our approach.

\subsection{Methodology}

In our study, we held interviews with four members of the FM JML, who each possess knowledge on adhesives but have varying degrees of involvement in adhesive selection.
Two interviewees are adhesive experts that often perform adhesive selection.
The other two do not perform adhesive selection as part of their job, but are knowledgeable on glues in general.
We included such ``non-experts'' to explore whether the tool is capable of making adhesive selection more accessible.
We held an online one-on-one session with each of the interviewees.
First the interviewee was asked to perform adhesive selection using the tool, and then we conducted a semi-structured interview to gather their opinions.

\paragraph{Adhesive selection.}
Two real-life bonding cases were selected by the second author to be used as test cases for the study: both are cases that the JML lab has received from companies in the past.
In the first one, a plastic door needs to be glued to the body of an industrial harvester.
Originally, the process of finding the correct glue took weeks, as the requirements are fairly tight.
Moreover, the specification contains an inconsistency in which a higher temperature is required than allowed by the substrate, which is the same inconsistency as shown in Fig.~\ref{fig:inconsistency}.
The second case details a join between a plastic component and an aluminum body, and is less challenging.

Both cases were presented to the participants as a short description of the gluing operation, together with a table containing the actual requirements.
For the first example, the table lists requirements such as ``Material~A~=~Virgin~ABS'', ``Gap~filling~of~min~1~mm'', ``Application~between~$15^\circ C$~to~$35^\circ C$'', etc.
We have taken care here to specify requirements in the same terminology as the tool.
In total, the use cases consist of nine and seven requirements respectively.

Before letting the experts work on the cases, we also gave a brief introduction on how to use the tool.
We explained how to enter information, where they could find the list of possible adhesives and how to see that the tool was performing calculations.

We observed each tester during the selection process, to gather information on how they interacted with the tool.
To better understand their thought process, the interviewees were encouraged to talk to themselves about their process as a way to elicitate their thoughts.
We did not intervene while they were working out these use cases, even when errors were made, and only made suggestions on what to do if they got stuck or an unexpected bug popped up.

\paragraph{Interviews.}
Right after the selection, we held an interview with each participant.
Because our goal is to explore the opinions of the JML members, we opted for a semi-structured interview set-up, for which we prepared four main questions to serve as a general guide:
\begin{enumerate}
    \item Do you see a role for the tool in your job?
    \item Do you feel that you understand what the tool does when you use it?
    \item What was your experience working with the tool?
    \item How would you compare our tool to the ones that you are used to working with?
\end{enumerate}
In addition to these main questions, we asked additional questions to zoom in on specific aspects of the answers given by the participants.
The interviews were led by the first author, with additional support from the second author.
They were audio-recorded and transcribed, so that they could be analysed thoroughly.

After transcribing the recordings, we followed the guidelines of \cite{Richards2018} and performed \textit{open coding} followed by \textit{axial coding} \cite[]{Corbin1990}.
Here, the goal is to identify various \textit{codes} that pop up during the interviews, and then further group them into several main categories.
These two steps were also performed separately by an external member of the research group, after which the results were compared and adapted to mitigate biases (\textit{consensus coding}).


\subsection{Results}\label{sss:interviews}


%
%
%

%



Based on the interview transcripts, we identified 26 codes in total.
Table~\ref{tab:stats} shows an overview of the interview statistics.
The last column of this table shows the \textit{Code distribution}, calculated as the cumulative percentage of codes discovered after each interview.
This is an important parameter that indicates whether \textit{data saturation} is reached, i.e., the point at which the same themes keep recurring and new interviews would not yield new results.
According to~\cite{Guest2020}, this point is reached when the difference in code distribution between the current and previous interview is $\leq 5\%$ (i.e., less than $5\%$ of new information was found).
As the information threshold between the third and fourth interview is 4\%, we conclude that we have reached data saturation.
A more detailed table showing the codes per interview is included in the Appendix.

\begin{table}
    \centering
    \caption{Interview statistics}
    \label{tab:stats}
    \begin{tabular}{ccccc}
        \topline
        Use Case Duration (min) & Interview Duration (min) & Words & Total codes & Code distribution
        \midline
        31 & 26 & 2092 & 15 & 58\%\\
        55 & 39 & 2107 & 16 & 88\%\\
        66 & 47 & 3497 & 19 & 96\%\\
        41 & 49 & 3680 & 13 & 100\%
        \botline
    \end{tabular}
\end{table}

\begin{figure}
    \includegraphics[width=\textwidth]{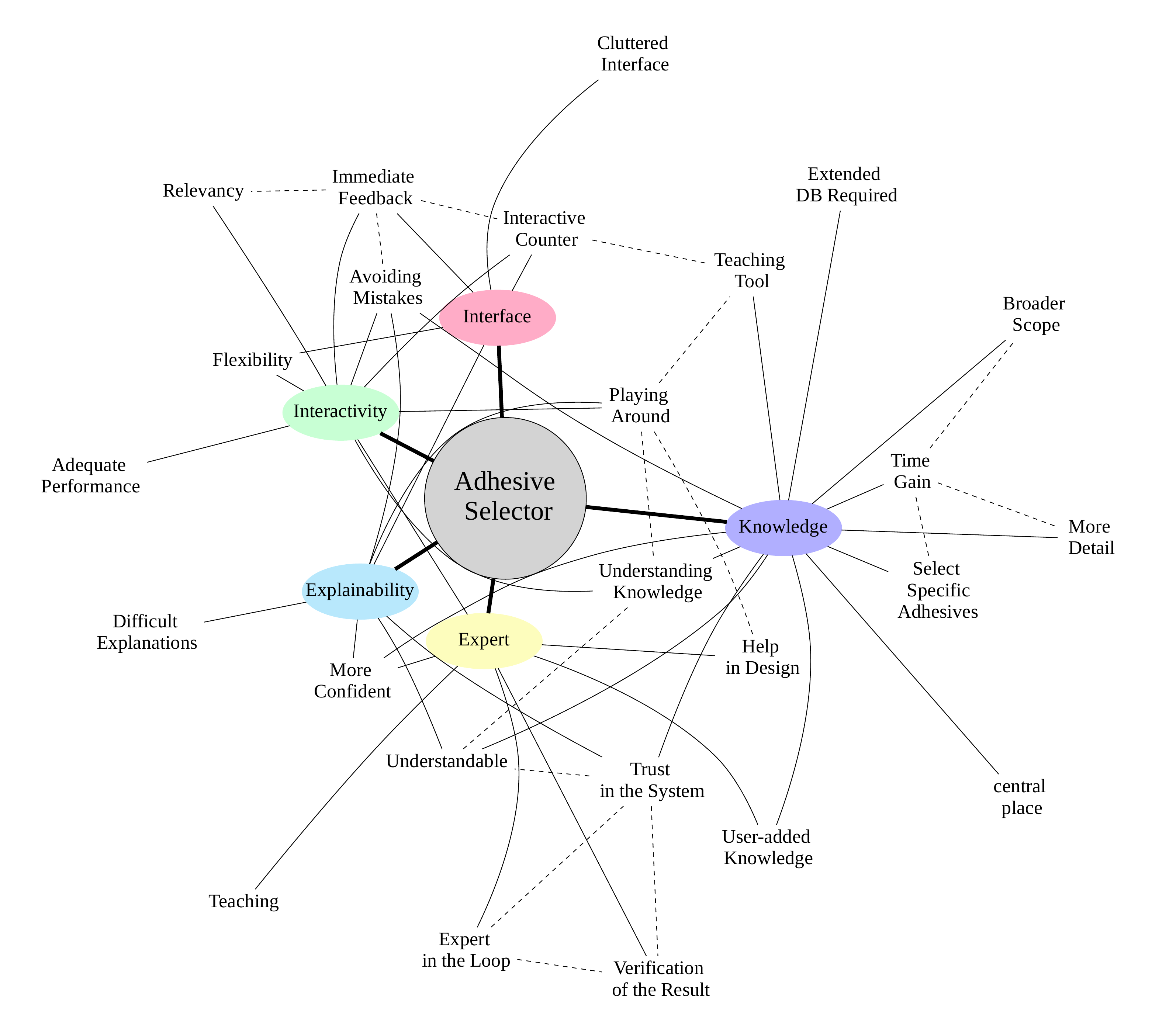}
    \caption{Graph showing connections between the interview themes and their codes.}
    \label{fig:mindmap}
\end{figure}

We sub-divided the codes into five main themes: Knowledge, Interactivity, Expert, Interface and Explainability.
These themes, together with their codes, are visualised in the graph in~Fig.~\ref{fig:mindmap}.
We will now briefly go over each theme and highlight the most important findings.

\subsubsection{Knowledge}
A major advantage of our tool is that it reasons on the knowledge of the adhesive experts ``like they themselves would''.
This is affirmed by the participants, who particularly liked its level of detail in two ways.
Firstly, instead of being limited to selecting adhesive families, as is the case in other selector systems, our tool can help them find specific adhesives.
Secondly, the number of parameters available in the tool is unparalleled, allowing for a more fine-grained search.

\blockquote{[...] the good thing that I like about this tool is that it’s quite detailed. It’s really one or multiple steps further, multiple steps deeper than these other tools. [...] So there it’s super helpful, definitely.}

Similarly, the tool can also efficiently reason on a broader number of specific adhesives than an expert.
As the latter typically knows some approximate parameter values by heart for a handful of glues and families, they tend to look at these adhesives first before widening their search to others when needed.
Here, the Adhesive Selector can help them to find additional adhesives that they would not have considered without the tool, while also saving them from having to manually go through their data sheets.

\blockquote{You might encounter situations in which you find an adhesive that you hadn't thought about before. That quite increases your search scope, I think. It could also help in mitigating bias.}

\blockquote{If you want to understand the glue that you want to use you have to read a lot of datasheets, while having this tool I think optimized the time in a way that is crazy.}

The main criticism expressed by the participants is that the number of specific adhesives in the knowledge base is still rather limited.
While 55 adhesives is a good start, adding more adhesives will definitely make the tool more effective.
Similarly, some participants remarked that the knowledge base should be with more expert knowledge such as the environmental impact and re-usability of adhesives.


\subsubsection{Interactivity}
Throughout the interviews, interactivity was an often recurring topic.
For example, one aspect that the participants all appreciated was that they could immediately see the effects of entering requirements in the interface.

\blockquote{Interactively choosing the glue now feels like online shopping -- I can select more options, and see the total number of adhesives go down until I have entered all requirements.}

Besides making the number of adhesives go down, each time a new requirement is entered the interface also greys out parameters that have become irrelevant, and removes parameter values that are no longer possible.
This immediate feedback helps preventing mistakes in the selection.

\blockquote{I think it’s easier to spot if there are some problems, like the ones that popped up.}
\blockquote{It’s safer, you avoid wasting materials and time.}

Moreover, the immediate feedback also allows the experts to ``play around'' with the knowledge in the tool.
In this way, they can get a feel for the effect of certain parameter values on the suitable adhesives.

\blockquote{Us engineers typically want to play with things. They want to see what happens when they change something, thereby implicitly performing a sensitivity analysis.}

However, some participants felt that always having to enter the requirements one-by-one is too inefficient.
While they all agreed that the tool is sufficiently fast, they stated that they would like to ``bulk update'' choices to be more efficient in cases where they already know that they are correct.

%
%

\subsubsection{Expert}
As our tool is designed to support JML's experts based on their own knowledge, they have played a big role in the creation of the tool and will continue to play a big role in its further existence.
It is necessary to always have an expert in the loop: the tool cannot be used by laymen, who lack the specific knowledge required to extract requirements from a description of a use case and who cannot understand the technical jargon.

Besides supporting the experts in making suitable selections, the participants described two other ways in which the Adhesive Selector can help them.
Firstly, the tool covers sufficient knowledge to also assist in the design of the entire joint.
This is different from adhesive selection in that many ``environmental parameters'' are still left open, e.g., substrates might need to be picked, a decision on joining method needs to be made, etc.

\blockquote{This tool can already help me to list all these requirements [required for the dimensioning of the joint] so that I have a bird’s eye view of the whole design.}

The second additional use of the tool identified by the participants is its potential to be used as a teaching tool.
For instance, newer members of the lab could use it to gain their footing when starting out.

\blockquote{You can use it to teach people `If you select these requirements, these are the consequences, which means you can no longer use these adhesives'.}
\blockquote{[The tool] can also give some confidence, if they say `I would select this' and the tool confirms it, you would feel more certain in your selection}

%

\subsubsection{Interface}
Having the participants talk out loud while working the use cases proved to be a valuable source of information, allowing us to gain insight in how someone without prior experience interacts with the tool.
Some of these insights are fairly minor, e.g., that we should order the values of the drop down lists alphabetically and support folding in/out entire categories to make the interface easier to navigate.
Other insights are more major, such as how much difficulty the participants experienced with the structure of the interface  -- due to the large number of parameter tiles (78 in total), it typically took them around 10 seconds to find the right one.
Moreover, the participants often lost overview of which choices they had already entered in the system, as these are spread all over the interface.

One positive note that the experts really liked was the automatically updating counter showing the remaining number of suitable adhesives:

\blockquote{I can look at the number of adhesives and see that we are converging, converging, converging. So it’s fun, to say it like that.}

This ties in nicely with the idea of the Adhesive Selector as a didactical tool: if a choice rules out many adhesives at once, the user can assume it is more ``important'' than a choice that only removes a handful.
Moreover, another participant suggested using diagrams to annotate symbols with their intended meaning, e.g., a diagram showing two bonded substrates with arrows pointing to the ``bond line'', the ``bonding surface'', etc.
This would make the tool more self-explanatory to people less familiar with these terms, such as the newer members of the team.

\subsubsection{Explainability}
While explainability is one of the focuses of our knowledge-based approach, the experts were not yet fully convinced of this functionality.
When prompted with the ``inconsistency'' window (Fig.~\ref{fig:inconsistency}), none of the participants knew what to do. 
Only two of them quickly understood what the cause of the inconsistency was, but none were able to resolve it by themselves.
As one participant later remarked in an interview:
\blockquote{There were multiple sentences below each other, I didn't know if it was three remarks or a  single one. [...] I was confused, and could not see the information I needed}
In other words, they had some difficulties navigating the inconsistency window: partly due to its layout, but also due to the complexity of the knowledge.
However, they did appreciate the potential that this feature holds, for example to assist in experimenting with the knowledge.

The difficult explanations also did not have an impact on the expert's trust in the system.
Indeed, it seems that it is more important to know that its behaviour is derived from the lab's own knowledge, than to actually 100\% understand the explanations.
\blockquote{I trust it, because it contains our knowledge. So, well, I do trust it, but only because I know it's ours.}

\paragraph{Other Results}

A participant with a slightly more business background pointed out an interesting result that could not be codified under the other themes.
They see the tool as a ``uniform way of collecting data relevant to adhesive selection'', which could help drive the team's decision making.
Examples of relevant information are the use cases performed by experts, the substrates that were used, which adhesives are typically picked, what constraints were present, etc.
This data could be useful for:
\begin{itemize}
    \item finding trainings specifically for the most commonly picked adhesives
    \item identifying target industries that could also be helped by JML
    \item deciding which adhesives to keep in stock, and what equipment should be purchased
\end{itemize}
This is a unique take that we had not yet considered for our tool.

\subsection{Limitations of the Study}

As with any study, ours is not without its limitations.
The main limitation of this study is our low number of interviewees.
However, as pointed out in Section~\ref{sss:interviews}, we do reach data saturation when looking at the code distribution of our interviews.
Therefore, we feel that the low number of interviewees does not have a major impact.

Another limitation is that all interviewees are from the same organization and would therefore have a ``common'' approach to selecting glues, while external experts might have a different focus that could result in additional feedback.
However, as the tool has specifically been developed for use within Flanders Make JML, we believe that this feedback would be less relevant.

During the testing of the tool, we continually observed the interviewees to study their interaction with the tool.
Because of this, the users might have felt pressured to work in a more ``efficient'' way, as a form of the Hawthorne effect~\cite[]{hawthorne}.
In fact, one participant said so explicitly, when talking about their experience with the inconsistency window:
\blockquote{I think that if I were alone, in the lab, I would have taken more time to read the pop-up. I wanted to be a bit quick.}
Because we were observing, we might have inhibited the users to truly ``play around'' with the tool and test it to their heart's content.

\section{Lessons Learned}\label{lessons}

Typically, knowledge acquisition is a time-consuming and difficult process.
We have found that the use of cDMN as a common notation can help facilitate this process.
The use of a formal representation that the experts can also understand helps to keep them in the loop and allows them to actively participate in the formalization process.
This way of working is less error-prone, as it functions as a preliminary validation of the knowledge.

After our three initial workshops, we mainly held one-on-one meetings with one of the experts to add information on the adhesives, and to further fine-tune the knowledge.
This resulted in a tight feedback loop, which turned out to be a key element in our formalization.
Indeed, thanks to thorough examinations of the tool by the expert, we were able to discover additional bugs in our KB.
Here, the Interactive Consultant was of paramount importance: each time the KB was modified, the expert could immediately play around with it using the generic interface.
In this way, the knowledge validation of the tool could happen immediately after the modifications, allowing for a swifter detection of any errors.

Having knowledge in a declarative format, independent of how it will be used, has multiple advantages.
To begin with, it allows using the knowledge for multiple purposes, even when this initially might not seem useful.
Furthermore, it increases the experts' trust in the system, as it reasons on the same knowledge as they do, and is interpretable.

The main advantage of using IDP-Z3 does not lie in any one of its inference algorithms, but rather in the fact that it allows all of the functionalities that are required for interactive exploration of the search space to be performed by applying different inference algorithms to a single KB. 

The validation of the tool by the actual end-users proved to be a source of valuable feedback.
Through our observations, we have gained insights on how the users interact with the tool.
By means of semi-structured interviews, we gathered their opinions, thoughts and suggestions.
These two sources of input combined will help shape further development of our tool.

\section{Conclusions and Future Work}\label{conclusion}
This paper presents the Adhesive Selector, a tool to support adhesive selection using a knowledge-based approach.
The Knowledge Base was constructed by conducting several workshops and one-on-one meetings, using a combination of DMN and cDMN.
Our current iteration of the tool contains sufficient knowledge to assist an expert in finding an initial list of adhesives.
Compared to the state of the art, it is declarative, more explainable, and more extensive.
The KB is also not limited to just adhesive selection, but can also be used to perform other related tasks.

In future work, we plan on converting the entire \fodot KB into cDMN, and evaluating its readability and maintainability from the perspective of the domain experts.
Besides this, we intend to test the tool using more real-life use cases, to quantify the gain in efficiency.
Additionally, we are also collaborating with an external research group to develop an AI-based tool capable of extracting adhesive parameter values from data sheets, to efficiently add more adhesives to our KB.
Flanders Make is planning on using the tool in production soon, and expanding the list of available adhesives is the only remaining bottle-neck.
Note that such information is only a part of the knowledge that is required for selecting a suitable adhesive, and other forms of expert knowledge will continue to be added throughout the lifetime of the tool through the methods described in the text.

\textbf{Competing interests:} The authors declare none.

\section*{Acknowledgements}

This research received funding from the Flemish Government under the ``Onderzoeksprogramma Artifici\"ele Intelligentie (AI) Vlaanderen'' programme and from Flanders Make vzw.
We would also like to thank Wouter Groeneveld for his suggestions on the interview structure, Benjamin Callewaert for cross-validating the codes and themes, and the members of the Joining and Materials Lab for participating in our study.

\section{Appendix}\label{appendix}

Table~\ref{tab:themes} shows an overview of each code and the interviews in which it appeared, and Table~\ref{tab:codebook} contains the ``code book'' explaining each of the codes.

While we cannot release the knowledge base itself, as it contains sensitive data from our external partner, the tools and techniques used in this paper are open-source and available online.
We refer anyone interested in trying out IDP-Z3\footnote{\url{https://www.IDP-Z3.be} and \url{https://gitlab.com/krr/IDP-Z3/}}, the Interactive Consultant\footnote{\url{https://interactive-consultant.idp-z3.be/}} or cDMN\footnote{\url{https://www.cdmn.be} and \url{https://gitlab.com/EAVISE/cdmn/cdmn-solver}} to their online resources.

\begin{table}[h!]
    \centering
    \caption{Discovered codes per participant, and the data saturation.}
    \label{tab:themes}
    \begin{tabular}{l l l l l}
        \topline
        Code & 1 & 2 & 3 & 4 
        \midline
        Cluttered Interface & \checkmark & \checkmark & \checkmark & \checkmark \\
        Interactive Counter & \checkmark & \checkmark & \checkmark &  \\
        Time Gain & \checkmark & \checkmark & \checkmark & \checkmark \\
        Avoiding Mistakes & \checkmark & & &\\
        Playing Around & \checkmark & & \checkmark &\\
        More Confident & \checkmark & & \checkmark & \checkmark\\
        Understandable & \checkmark & & & \\
        Trust in the System & \checkmark & \checkmark & \checkmark & \checkmark \\
        Verification of the Result & \checkmark & \checkmark & \checkmark & \checkmark \\
        Adequate Performance & \checkmark & \checkmark & \checkmark & \checkmark \\
        Broader Scope & \checkmark &  & \checkmark & \checkmark \\
        Understanding Knowledge & \checkmark &  &  & \checkmark \\
        Teaching Tool & \checkmark &  & \checkmark & \checkmark \\
        Explainability & \checkmark &  \checkmark & \checkmark &\\
        Difficult Explanations & \checkmark &  \checkmark & \checkmark & \checkmark \\
        Extended DB Required &  &  \checkmark & \checkmark &  \\
        Select Specific Adhesives &  &  \checkmark & \checkmark &  \\
        Help in Design &  &  \checkmark &  &  \\
        Expert in the Loop &  &  \checkmark &  \checkmark & \checkmark \\
        Interactivity &  &  \checkmark &  \checkmark & \\
        Relevancy &  &  \checkmark &  & \\
        More Detail &  &  \checkmark & \checkmark & \checkmark \\
        User-added Knowledge &  &  \checkmark & & \\
        Flexibility &  &  & \checkmark & \\
        Immediate Feedback &  &  & \checkmark & \\
        Central Place for Knowledge &  &  & & \checkmark 
        \midline
        Total codes & 15 & 16 & 19 & 13 \\
        Data saturation & 58\% & 88\% & 96\% & 100\% 
        \botline
    \end{tabular}
\end{table}

\begin{table}[h!]
    \centering
    \begin{tabular}{ll}
        \topline
        Code & Explanation
        \midline
        Cluttered Interface & The interface's structure seems chaotic. \\
        Interactive Counter & The counter denoting the number of suitable adhesives.\\
        Time Gain & Experts can more efficiently select an adhesive.\\
        Avoiding Mistakes & The tool prevents making mistakes.\\
        Playing Around & \makecell[l]{Interacting with the knowledge allows the expert to \\``play around'' with it.}\\
        More Confident & The experts are more sure of their selection using the tool.\\
        Understandable & \makecell[l]{The interface presents information in a sufficiently\\ understandable way}\\
        Trust in the System & The experts trust the choices of the system.\\
        Verification of the Result & \makecell[l]{The resulting short-list of adhesives will always need to be\\ verified.}\\
        Adequate Performance & The loading time of the tool is within reason.\\
        Broader Scope & The experts discover more adhesives.\\
        Understanding Knowledge & The tool helps understand the knowledge.\\
        Teaching Tool & The tool is well-suited for teaching newer members of the lab.\\
        Explainability & Explainability is an important factor.\\
        Difficult Explanations & The automatically generated explanations are difficult.\\
        Extended DB Required & More adhesives should be added to the system.\\
        Select Specific Adhesives & \makecell[l]{The AS allows selecting specific adhesives, whereas the other\\ tools only select families.}\\
        Help in Design & The tool can also play a role in joint design.\\
        Expert in the Loop & An expert is always required to stay in the loop.\\
        Interactivity & Interactivity is a crucial aspect of the tool.\\
        Relevancy & \makecell[l]{Detecting which parameters are still relevant aids in selecting\\ adhesives.}\\
        More Detail & \makecell[l]{The tool is more detailed than any other in terms of possible\\ parameters.}\\
        User-added Knowledge & \makecell[l]{The experts would like to add their own knowledge to the\\ system.}\\
        Flexibility & The tool adapts well to the expert's way of working.\\
        Immediate Feedback &  \makecell[l]{The experts appreciate that the interface updates\\ automatically.}\\
        Central Place for Knowledge & \makecell[l]{All knowledge for adhesive selection is stored in one\\ location.}
        \botline
    \end{tabular}
    \caption{A ``code book'' elaborating on the meaning of each code.}
    \label{tab:codebook}
\end{table}

\bibliographystyle{tlplike}
\bibliography{biblio}

\end{document}